\documentclass[table]{article}
\pdfoutput=1

\usepackage{microtype}
\usepackage{graphicx}
\usepackage{booktabs} 
\usepackage{amsmath}
\usepackage{amssymb}
\usepackage{pgfplots} 
\usepackage{colortbl}
\usepackage{afterpage}

\usepackage{subfig} 
\usepackage{caption}
\usepackage{lipsum}
\usepackage[colorinlistoftodos, textsize=tiny]{todonotes}
\definecolor{citrine}{rgb}{0.89, 0.82, 0.04}
\definecolor{blued}{RGB}{70,197,221}

\pgfplotsset{
	tick label style={font=\footnotesize},
	label style={font=\small},
	legend style={font=\small},
}

\usepackage{listings}
\usepackage{hyperref}
\usepackage[capitalize,noabbrev]{cleveref}

\usepackage[accepted]{icml2019}

\icmltitlerunning{GANs for Anomaly Detection: a survey}

\begin{document}

\twocolumn[
	\icmltitle{A Survey on GANs for Anomaly Detection}



	\icmlsetsymbol{equal}{*}

	\begin{icmlauthorlist}
		\icmlauthor{Federico Di Mattia}{to,equal}
		\icmlauthor{Paolo Galeone}{to,equal}
		\icmlauthor{Michele De Simoni}{to,equal}
		\icmlauthor{Emanuele Ghelfi}{to,equal}
	\end{icmlauthorlist}

	\icmlaffiliation{to}{Zuru Tech, Modena, Italy}

    \icmlcorrespondingauthor{Federico Di Mattia}{federico.d@zuru.tech}

	\icmlkeywords{Machine Learning, ICML, Deep Learning, Anomaly Detection, GANs}

	\vskip 0.3in
]



\printAffiliationsAndNotice{\icmlEqualContribution} 

\begin{abstract}
	Anomaly detection is a significant problem faced in several research areas. Detecting and correctly classifying something unseen as anomalous is a challenging problem that has been tackled in many different manners over the years.
	Generative Adversarial Networks (GANs) and the adversarial training process have been recently employed to face this task yielding remarkable results. In this paper we survey the principal GAN-based anomaly detection methods, highlighting their pros and cons. Our contributions are the empirical validation of the main GAN models for anomaly detection, the increase of the experimental results on different datasets and the public release of a complete Open Source toolbox for Anomaly Detection using GANs.
\end{abstract}

\section{Introduction}
\label{sec:intro}

Anomalies are patterns in data that do not conform to a well-defined notion of normal behavior \cite{Chandola2009}.
Generative Adversarial Networks (GANs) and the adversarial training framework \cite{Goodfellow2014} have been successfully applied to model complex and high dimensional distribution of real-world data. This GAN characteristic suggests they can be used successfully for anomaly detection, although their application has been only recently explored.
Anomaly detection using GANs is the task of modeling the normal behavior using the adversarial training process and detecting the anomalies measuring an \textit{anomaly score} \cite{Schlegl2017}.
To the best of our knowledge, all the GAN-based approaches to anomaly detection build upon on the Adversarial Feature Learning idea \cite{Donahue2016} in which the BiGAN architecture has been proposed.
In their original formulation, the GAN framework learns a generator that maps samples from an arbitrary latent distribution (noise prior) to data as well as a discriminator which tries to distinguish between real and generated samples.
The BiGAN architecture extended the original formulation, adding the learning of the \textit{inverse mapping} which maps the data back to the latent representation.
A learned function that maps input data to its latent representation together with a function that does the opposite (the generator) is the basis of the anomaly detection using GANs.

The paper is organized as follows. In \cref{sec:intro} we introduce the GANs framework and, briefly, its most innovative extensions, namely conditional GANs and BiGAN, respectively in \cref{sec:cgan} and \cref{sec:bigan}. \cref{sec:gan-anomaly} contains the state of the art architectures for anomaly detection with GANs. In \cref{sec:experiments} we empirically evaluate all the analyzed architectures. Finally, \cref{sec:conclusions} contains the conclusions and future research directions.

\subsection{GANs}
\label{sec:gan}

GANs are a framework for the estimation of generative models via an adversarial process in which two models, a discriminator $D$ and a generator $G$, are trained simultaneously.
The generator $G$ aim is to capture the data distribution, while the discriminator $D$ estimates the probability that a sample came from the training data rather than $G$.
To learn a generative distribution $p_g$ over the data $\mathbf{x}$ the generator builds a mapping from a prior noise distribution $p_z$ to a data space as $G(\mathbf{z};\theta_G)$, where $\theta_G$ are the generator parameters.
The discriminator outputs a single scalar representing the probability that $\mathbf{x}$ came from real data rather than from $p_g$. The generator function is denoted with $D(\mathbf{x}; \theta_D)$, where $\theta_D$ are discriminator parameters.

The original GAN framework \cite{Goodfellow2014} poses this problem as a min-max game in which the two players ($G$ and $D$) compete against each other, playing the following zero-sum min-max game:

\vspace{-0.5cm}
\begin{equation}
	\begin{aligned}
		\min_G \max_D & \; V(D,G) = \mathbb{E} _{\mathbf{x} \sim p_{data}(\mathbf{x})}[\log D(\mathbf{x})] +                 \\
		              & \mathbb{E}_{\mathbf{z} \sim p_z(\mathbf{z})}\left[ \log \left( 1 - D(G(\mathbf{z})) \right) \right].
	\end{aligned}
\end{equation}
\begin{figure*}[t!]
	\centering
	\includegraphics[width=0.7\textwidth]{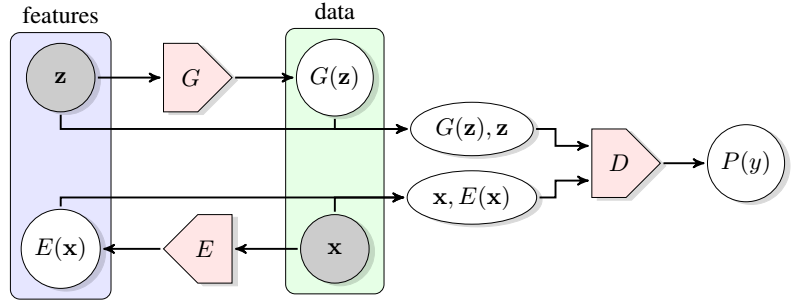}
	\caption{The structure of BiGAN proposed in \cite{Donahue2016}.}
	\label{fig:bigan_structure}
\end{figure*}
\subsection{Conditional GANs}
\label{sec:cgan}

GANs can be extended to a conditional model \cite{Mirza2014} conditioning either $G$ or $D$ on some extra information $y$. The $y$ condition could be any auxiliary information, such as class labels or data from other modalities. We can perform the conditioning by feeding $y$ into both the discriminator and generator as an additional input layer.
The generator combines the noise prior $p_z(\mathbf{z})$ and $y$ in a joint hidden representation, the adversarial training framework allows for considerable flexibility in how this hidden representation is composed.
In the discriminator, $\mathbf{x}$ and $y$ are presented as inputs to a discriminative function.
The objective function considering the condition becomes:

\begin{equation}
	\begin{aligned}
		\min_G \max_D & \; V(D,G) = \mathbb{E} _{\mathbf{x} \sim p_{data}(\mathbf{x}|y)}[\log D(\mathbf{x})] + \\
		              & \mathbb{E}_{\mathbf{z} \sim p_z(\mathbf{z})}[\log(1 - D(G(\mathbf{z}|y)))].
	\end{aligned}
\end{equation}

\subsection{BiGAN}
\label{sec:bigan}

Bidirectional GAN \cite{Donahue2016} extends the GAN framework including an encoder $E(x; \theta_E)$ that learns the inverse of the generator $E = G^{-1}$. The BiGAN training process allows learning a mapping simultaneously from latent space to data and vice versa. The encoder $E$ is a non-linear parametric function in the same way as $G$ and $D$, and it can be trained using gradient descent.
As in the conditional GANs scenario, the discriminator must learn to classify not only real and fake samples, but pairs in the form $(G(\mathbf{z}), \mathbf{z})$ or $(\mathbf{x}, E(\mathbf{x}))$. The BiGAN training objective is:

\begin{equation}
	\label{BiGAN_equation}
	\begin{aligned}
		\min_{G,E} \max_D & \; V(D,G,E) =                                                                                                                                    \\
		                  & \mathbb{E}_{ \mathbf{x} \sim p_{data}(\mathbf{x})}[\mathbb{E}_{\mathbf{z} \sim p_{E(\mathbf{z}|\mathbf{x})}} [\log D(\mathbf{x}, \mathbf{z})]] + \\
		                  & \mathbb{E}_{\mathbf{z} \sim p_z(\mathbf{z})}[\mathbb{E}_{\mathbf{x} \sim p_G(\mathbf{x}|\mathbf{z})}[\log(1 - D(\mathbf{x},\mathbf{z})))]].
	\end{aligned}
\end{equation}

\cref{fig:bigan_structure} depicts a visual structure of the BiGAN architecture.

\section{GANs for anomaly detection}
\label{sec:gan-anomaly}

Anomaly detection using GANs is an emerging research field. \citet{Schlegl2017}, here referred to as AnoGAN, were the first to propose such a concept. In order to face the performance issues of AnoGAN a BiGAN-based approach has been proposed in \citet{Zenati2018}, here referred as EGBAD (Efficient GAN Based Anomaly Detection), that outperformed AnoGAN execution time.
Recently, \citet{Akcay2018} advanced a GAN + autoencoder based approach that exceeded EGBAD performance from both evaluation metrics and execution speed.

In the following sections, we present an analysis of the considered architecture. The term sample and image are used interchangeably since GANs can be used to detect anomalies on a wide range of domains, but all the analyzed architectures focused mostly on images.

\subsection{AnoGAN}
\label{sec:anogan}

\begin{figure*}[t!]
	\centering
	\includegraphics[width=0.97\textwidth]{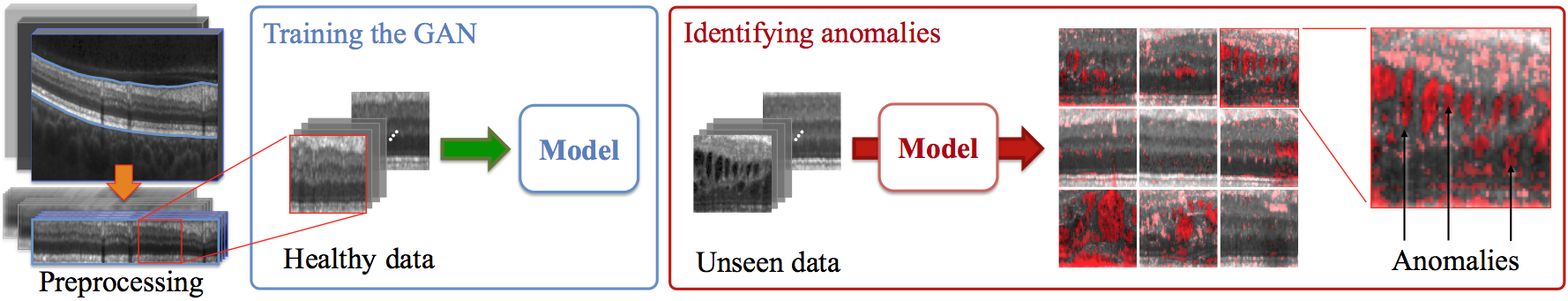}
	\caption{AnoGAN \cite{Schlegl2017}. The GAN is trained on positive samples. At test time, after $\Gamma$ research iteration the latent vector that maps the test image to its latent representation is found $\mathbf{z}_\Gamma$. The reconstructed image $G(\mathbf{z}_{\Gamma})$ is used to localize the anomalous regions.}
	\label{fig:anogan_framework}
\end{figure*}

AnoGAN aim is to use a standard GAN, trained only on positive samples, to learn a mapping from the latent space representation $\mathbf{z}$ to the realistic sample $\hat{\mathbf{x}} = G(\mathbf{z})$ and use this learned representation to map new, unseen, samples back to the latent space.
Training a GAN on normal samples only, makes the generator learn the manifold $\mathcal{X}$ of normal samples. Given that the generator learns how to generate normal samples, when an anomalous image is encoded its reconstruction will be non-anomalous; hence the difference between the input and the reconstructed image will highlight the anomalies. The two steps of training and detecting anomalies are summarized in \cref{fig:anogan_framework}.

The authors have defined the mapping of input samples to the latent space as an iterative process. The aim is to find a point $\mathbf{z}$ in the latent space that corresponds to a generated value $G(\mathbf{z})$ that is similar to the query value $\mathbf{x}$ located on the manifold $\mathcal{X}$ of the positive samples.
The research process is defined as the minimization trough $\gamma = 1,2, \dots, \Gamma$ backpropagation steps of the loss function defined as the weighted sum of the residual loss $\mathcal{L}_R$ and discriminator loss $\mathcal{L}_D$, in the spirit of \citet{Yeh2016}.

The \textit{residual loss} measures the dissimilarity between the query sample and the generated sample in the input domain space:

\vspace{-1em}

\begin{equation}
	\mathcal{L}_{R}(\mathbf{z}_\gamma) = ||\mathbf{x} - G(\mathbf{z}_\gamma)||_1.
	\label{residual_loss}
\end{equation}

The \textit{discriminator loss} takes into account the discriminator response. It can be formalized in two different ways. Following the original idea of \citet{Yeh2016}, hence feeding the generated image $G(\mathbf{z}_\gamma)$ into the discriminator and calculating the sigmoid cross-entropy as in the adversarial training phase: this takes into account the discriminator confidence that the input sample is derived by the real data distribution. Alternatively, using the idea introduced by \citet{Salimans2016}, and used by the AnoGAN \cite{Schlegl2017} authors, to compute the feature matching loss, extracting features from a discriminator layer $\mathbf{f}$ in order to take into account if the generated sample has similar features of the input one, by computing:
\vspace{-1.5em}

\begin{align}
	\mathcal{L}_{D}(\mathbf{z}_\gamma) = ||\mathbf{f}(\mathbf{x}) - \mathbf{f}(G(\mathbf{z}_\gamma))||_1,
\end{align}

hence the proposed loss function is:
\vspace{-1.5em}

\begin{align}
	\mathcal{L}(\mathbf{z}_\gamma) = (1 - \lambda) \cdot \mathcal{L}_{R}(\mathbf{z}_\gamma) + \gamma \cdot \mathcal{L}_D(\mathbf{z}_\gamma).
\end{align}

Its value at the $\Gamma$-th step coincides with the \textit{anomaly score} formulation:
\begin{align}
	A(\mathbf{x}) = \mathcal{L}(\mathbf{z}_\Gamma).
\end{align}

$A(\mathbf{x})$ has no upper bound; to high values correspond an high probability of $\mathbf{x}$ to be anomalous.

It should be noted that the minimization process is required for every single input sample $\mathbf{x}$.

\subsubsection{Pros and cons}
\textbf{Pros}
\begin{itemize}
	\item Showed that GANs can be used for anomaly detection.
	\item Introduced a new mapping scheme from latent space to input data space.
	\item Used the same mapping scheme to define an anomaly score.
\end{itemize}
\textbf{Cons}
\begin{itemize}
	\item Requires $\Gamma$ optimization steps for every new input: bad test-time performance.
	\item The GAN objective has not been modified to take into account the need for the inverse mapping learning.
	\item The anomaly score is difficult to interpret, not being in the probability range.
\end{itemize}

\subsection{EGBAD}
\label{sec:egbad}
Efficient GAN-Based Anomaly Detection (EGBAD) \cite{Zenati2018} brings the BiGAN architecture to the anomaly detection domain. In particular, EGBAD tries to solve the AnoGAN disadvantages using \citet{Donahue2016} and \citet{Dumoulin2017} works that allows learning an encoder $E$ able to map input samples to their latent representation during the adversarial training. The importance of learning $E$ jointly with $G$ is strongly emphasized, hence \citet{Zenati2018} adopted a strategy similar to the one indicated in \citet{Donahue2016} and \citet{Dumoulin2017} in order to try to solve, during training, the optimization problem $\min_{G,E} \max_{D} V(D, E, G)$ where $V(D, E, G)$ is defined as in Equation \ref{BiGAN_equation}.
The main contribution of the EGBAD is to allow computing the anomaly score without $\Gamma$ optimization steps during the inference as it happens in AnoGAN \cite{Schlegl2017}.

\subsection{GANomaly}
\label{sec:ganomaly}

\begin{figure*}[t!]
	\centering
	\includegraphics[width=0.8\textwidth]{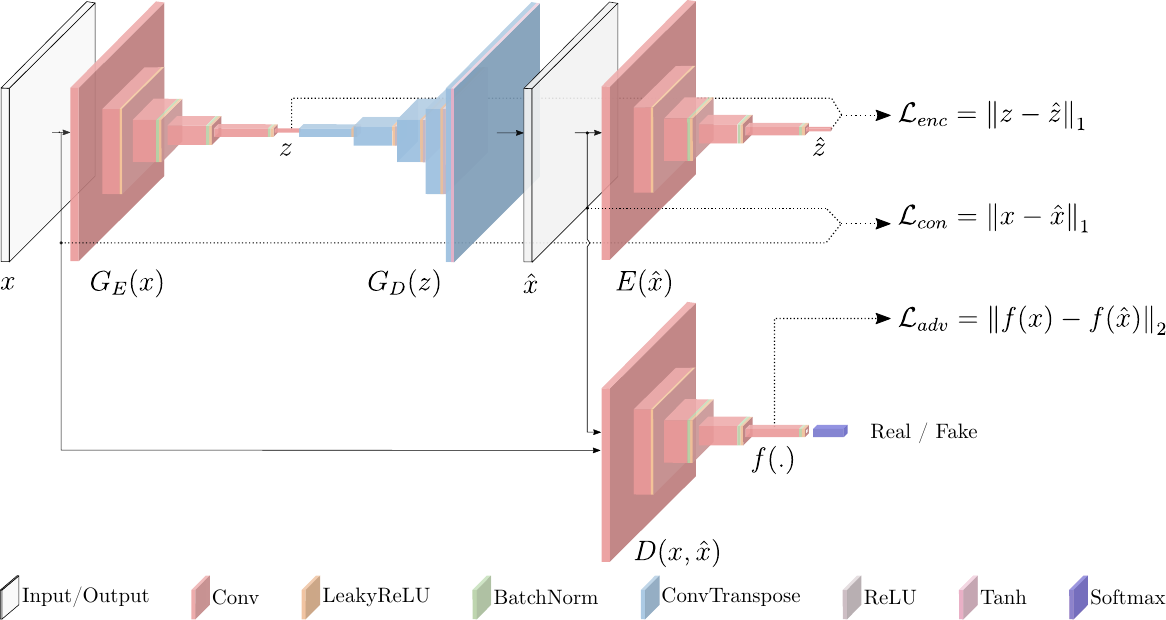}
	\caption{GANomaly architecture and loss functions from \cite{Akcay2018}.}
	\label{fig:ganomaly_pipeline}
\end{figure*}

\citet{Akcay2018} introduce the GANomaly approach. Inspired by AnoGAN \cite{Schlegl2017}, BiGAN \cite{Donahue2016} and EGBAD \cite{Zenati2018} they train a generator network on normal samples to learn their manifold $\mathcal{X}$ while at the same time an autoencoder is trained to learn how to encode the images in their latent representation efficiently. Their work is intended to improve the ideas of \citet{Schlegl2017}, \citet{Donahue2016} and \citet{Zenati2018}.
Their approach only needs a generator and a discriminator as in a standard GAN architecture.

\textbf{Generator network} The generator network consists of three elements in series, an encoder $G_E$ a decoder $G_D$ (both assembling an autoencoder structure) and another encoder $E$. The architecture of the two encoders is the same. $G_E$ takes in input an image $\mathbf{x}$ and outputs an encoded version $\mathbf{z}$ of it. Hence, $\mathbf{z}$ is the input of $G_D$ that outputs $\mathbf{\hat{x}}$, the reconstructed version of $\mathbf{x}$. Finally, $\mathbf{\hat{x}}$ is given as an input to the encoder $E$ that produces $\mathbf{\hat{z}}$. There are two main contributions from this architecture.
First, the operating principle of the anomaly detection of this work lies in the autoencoder structure. Given that we learn to encode normal (non-anomalous) data (producing $\mathbf{z}$) and given that we learn to generate normal data ($\mathbf{\hat{x}}$) starting from the encoded representation $\mathbf{z}$, when the input data $\mathbf{x}$ is an anomaly its reconstruction will be normal. Because the generator will always produce a non-anomalous image, the visual difference between the input $\mathbf{x}$ and the produced $\mathbf{\hat{x}}$ will be high and in particular will spatially highlight where the anomalies are located.
Second, the encoder $E$ at the end of the generator structure helps, during the training phase, to learn to encode the images in order to have the best possible representation of $\mathbf{x}$ that could lead to its reconstruction $\mathbf{\hat{x}}$.

\textbf{Discriminator network} The discriminator network $D$ is the other part of the whole architecture, and it is, with the generator part, the other building block of the standard GAN architecture. The discriminator, in the standard adversarial training, is trained to discern between real and generated data. When it is not able to discern among them, it means that the generator produces realistic images. The generator is continuously updated to fool the discriminator.
Refer to \cref{fig:ganomaly_pipeline} for a visual representation of the architecture underpinning GANomaly.

\begin{figure*}[t!]
	\centering
	\includegraphics[width=0.97\textwidth]{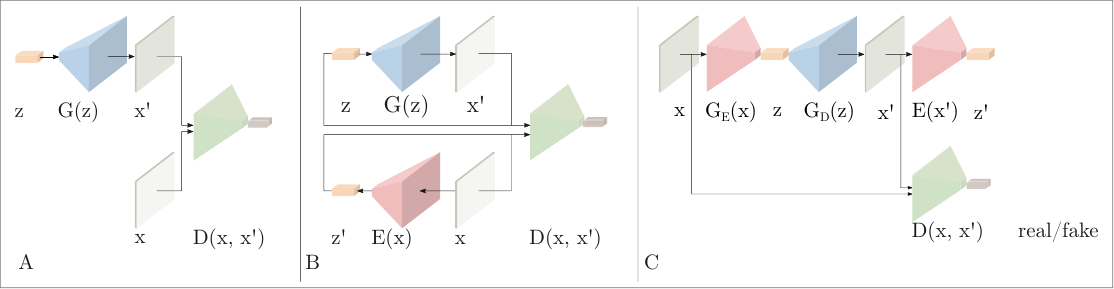}
	\caption{Architectures comparison. A) AnoGAN \cite{Schlegl2017}, B) EGBAD \cite{Zenati2018}, C) GANomaly \cite{Akcay2018}.}
	\label{fig:ganomaly_comparison}
\end{figure*}

The GANomaly architecture differs from AnoGAN \cite{Schlegl2017} and from EGBAD \cite{Zenati2018}. In \cref{fig:ganomaly_comparison} the three architectures are presented.

Beside these two networks, the other main contribution of GANomaly is the introduction of the generator loss as the sum of three different losses; the discriminator loss is the classical discriminator GAN loss.

\textbf{Generator loss} The objective function is formulated by combining three loss functions, each of which optimizes a different part of the whole architecture.

\textit{Adversarial Loss} The adversarial loss it is chosen to be the feature matching loss as introduced in \citet{Schlegl2017} and pursued in \citet{Zenati2018}:

\begin{equation}
	\begin{aligned}
		\mathcal{L}_{adv} = \mathbb{E}_{\mathbf{x} \sim p_{X}}||f(\mathbf{x})-\mathbb{E}_{\mathbf{x} \sim p_{X}} f(G(\mathbf{x}))||_2,
	\end{aligned}
\end{equation}

where $f$ is a layer of the discriminator $D$, used to extract a feature representation of the input .
Alternatively, binary cross entropy loss can be used.

\textit{Contextual Loss} Through the use of this loss the generator learns contextual information about the input data. As shown in \cite{Isola2016} the use of the $\mathcal{L}_{1}$ norm helps to obtain better visual results:

\begin{equation}
	\begin{aligned}
		\mathcal{L}_{con} = \mathbb{E}_{\mathbf{x} \sim p_{X}}||x-G(\mathbf{x})||_1.
	\end{aligned}
\end{equation}

\textit{Encoder Loss} This loss is used to let the generator network learn how to best encode a normal (non-anomalous) image:

\begin{equation}
	\begin{aligned}
		\mathcal{L}_{enc} = \mathbb{E}_{\mathbf{x} \sim p_{X}}||G_E(\mathbf{x})-E(G(\mathbf{x}))||_2.
	\end{aligned}
\end{equation}

The resulting generator loss will be the result of the (weighted) sum of the three previously defined losses:

\begin{equation}
	\begin{aligned}
		\mathcal{L} = w_{adv} \mathcal{L}_{adv} + w_{con} \mathcal{L}_{con} + w_{enc} \mathcal{L}_{enc} \; ,
	\end{aligned}
\end{equation}

where $w_{adv}$, $w_{con}$ and $w_{enc}$ are weighting parameters used to adjust the importance of the three losses.

At test stage, the authors proposed to compute the anomaly score in a different way respect to AnoGAN: only using using $\mathcal{L}_{enc}$:

\begin{equation}
	\begin{aligned}
		\mathcal{A}(\mathbf{x}) = ||G_E(\mathbf{x})-E(G(\mathbf{x}))||_2 \; .
	\end{aligned}
\end{equation}

In order to make the anomaly score easier to interpret, they proposed to compute the anomaly score for every sample $\mathbf{\hat{x}}$ in the test set $\hat{D}$, getting a set of individual anomaly scores: $\mathcal{S} = \left\{ s_i : \mathcal{A}(\hat{x}_i) \,, \hat{x}_i \in \hat{D} \right\}$ and then apply feature scaling to have the anomaly scores within the probabilistic range of $[0,1]$:

\begin{equation}
	\begin{aligned}
		s'_{i} = \frac{s_i + \min(S)}{\max(S) - \min(S)} \; .
	\end{aligned}
\end{equation}

\subsubsection{Pros and cons}
\textbf{Pros}
\begin{itemize}
	\item An encoder is learned during the training process, hence we don't have the need for a research process as in AnoGAN \cite{Schlegl2017}.
	\item Using an autoencoder like architecture (no use of noise prior) makes the entire learning process faster.
	\item The anomaly score is easier to interpret.
	\item The contextual loss can be used to localize the anomaly.
\end{itemize}
\textbf{Cons}
\begin{itemize}
	\item It allows to detect anomalies both in the image space and in the latent space, but the results couldn't match: a higher anomaly score, that's computed only in the latent space, can be associated with a generated sample with a low contextual loss value and thus very similar to the input - and vice versa.
	\item Defines a new anomaly score.
\end{itemize}

\section{Experiments}
\label{sec:experiments}

To evaluate the performance of every Anomaly Detection algorithm described we re-implemented all of them using the popular deep learning framework Tensorflow \cite{Tensorflow2015} creating a publicly available toolbox for the Anomaly Detection with GANs.

The toolbox is available on PyPI\footnote{\url{https://pypi.org/project/anomaly-toolbox/}}, which you can install using the install command:
\begin{lstlisting}
  pip install anomaly-toolbox
\end{lstlisting}
or, alternatively, on GitHub\footnote{\url{https://github.com/zurutech/anomaly-toolbox}} if you want to collaborate and explore the code.

Experiments were made trying to improve, where possible (i.e., where there were known errors communicated directly by the authors), the code. The results shown in the following sections are the best empirically obtained among all the tests carried out and do not necessarily derive from the configurations of the networks that at a theoretical level should be the correct ones.
In particular, as described in \cref{sec:appendix}, in version 1 (\cref{sec:version_1}) of our tests, with the correct implementation of the BiGAN/EGBAD models, we obtained worse results than those obtained in version 2 (\cref{sec:version_2}) of our tests with the architecture already implemented in the reference paper, although this contained known errors.

\begin{figure*}[ht]
	\centering
	{\centering \textbf{BiGAN/EGBAD and GANomaly performance comparison on MNIST and Fashion-MNIST datasets}\par\medskip}
	\begin{tikzpicture}[scale=1,line width=1pt]

		\begin{axis}[
				xlabel={(a) MNIST anomalous classes},
				ylabel={AUPRC},
				xmin=-0.5, xmax=9.5,
				ymin=0, ymax=1.1,
				xtick={0, 1, 2, 3, 4, 5, 6, 7, 8, 9},
				ytick={0.2, 0.4, 0.6, 0.8, 1},
				legend pos=north west,
				legend style={font=\fontsize{4}{5}\selectfont},
				ymajorgrids=true,
				grid style=dashed,
			]
			\addplot[ 
				only marks,
				color=blue,
				mark=triangle,
				error bars/.cd, y dir=both,y explicit
			]
			coordinates {
					(0,0.836239)
					(1,0.873284)
					(2,0.879434)
					(3,0.787722)
					(4,0.746983)
					(5,0.788937)
					(6,0.842500)
					(7,0.827540)
					(8,0.778773)
					(9,0.554513)
				};

			\addplot[ 
				only marks,
				color=green,
				mark=*,
				error bars/.cd, y dir=both,y explicit
			]
			coordinates {
					(0,0.667249)
					(1,0.248307)
					(2,0.732796)
					(3,0.568800)
					(4,0.568901)
					(5,0.603013)
					(6,0.669514)
					(7,0.401656)
					(8,0.702614)
					(9,0.398312)
				};

			\legend{BiGAN/EGBAD, GANomaly}
		\end{axis}
	\end{tikzpicture}
	\begin{tikzpicture}[scale=1,line width=1pt]
		\begin{axis}[
				xlabel={(b) Fashion-MNIST anomalous classes},
				ylabel={AUPRC},
				xmin=-0.5, xmax=9.5,
				ymin=0, ymax=1.1,
				xtick={0, 1, 2, 3, 4, 5, 6, 7, 8, 9},
				ytick={0.2, 0.4, 0.6, 0.8, 1},
				legend pos=north west,
				legend style={font=\fontsize{4}{5}\selectfont},
				ymajorgrids=true,
				grid style=dashed,
			]
			\addplot[ 
				only marks,
				color=blue,
				mark=triangle,
				error bars/.cd, y dir=both,y explicit
			]
			coordinates {
					(0,0.682896)
					(1,0.917813)
					(2,0.686013)
					(3,0.713173)
					(4,0.594117)
					(5,0.708108)
					(6,0.594738)
					(7,0.848828)
					(8,0.886060)
					(9,0.874649)

				};

			\addplot[ 
				only marks,
				color=green,
				mark=*,
				error bars/.cd, y dir=both,y explicit
			]
			coordinates {
					(0,0.410541)
					(1,0.789967)
					(2,0.385871)
					(3,0.485164)
					(4,0.424755)
					(5,0.837918)
					(6,0.403830)
					(7,0.528115)
					(8,0.868232)
					(9,0.637121)
				};

			\legend{BiGAN/EGBAD, GANomaly}
		\end{axis}
	\end{tikzpicture}
	\caption{Performance on MNIST (a) and Fashion-MNIST (b) of BiGAN/EGBAD and GANomaly models measured by the area under the precision-recall curve (AUPRC). The results have been selected as the best result between different type of training (with bce/fm and with residual loss). Best viewed in color. }
	\label{img:BiGAN_GANomaly_resutls}
\end{figure*}
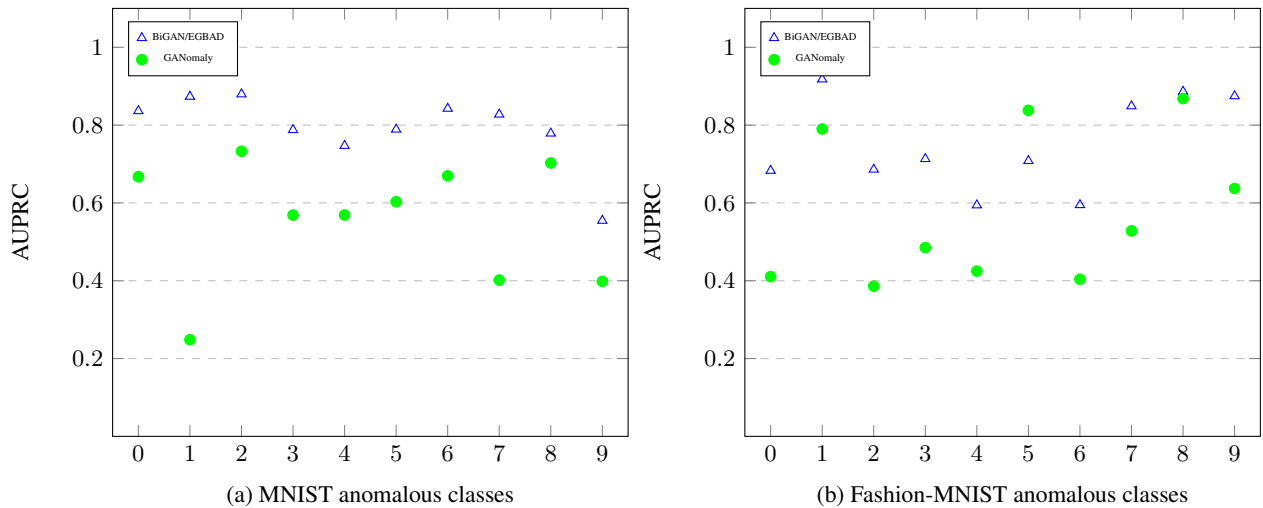
\subsection{Experimental setup}

The experiments are performed using a Intel\textsuperscript{\textregistered} Core\textsuperscript{\texttrademark} i7-7820X CPU and NVIDIA\textsuperscript{\textregistered} GTX 1080 Ti.

\subsubsection{Datasets}

We decided to train and test on widely known datasets commonly found in the literature: MNIST \cite{lecun-mnisthandwrittendigit-2010}, FashionMNIST \cite{Xiao2018}, CIFAR-10 \cite{cifar10}, and KDD99 \cite{Lichman2013}.

\textbf{MNIST:}
MNIST dataset \cite{lecun2010} is a database of handwritten digits. It consists of $28 \times 28$ pixels grayscale images split in a training set of 60,000 examples and a test set of 10,000 examples. The output classes are 10 in total and represent the ten digits from 0 to 9. In order to replicate the results of the papers presented in this work, the training process and the subsequent testing phase have been evaluated on each class, i.e., one at a time, only one class has been considered as the anomaly class, and the remaining nine classes have been considered together as the normal data.

\textbf{Fashion MNIST:}
Fashion-MNIST dataset \cite{Xiao2018} is a dataset intended to be a more complex drop-in replacement for the traditional MNIST dataset, developed by Zalando\textsuperscript{\textregistered} Research Group. It shares with the MNIST dataset the same image size and train-test split; it consists of $28 \times 28$ pixels images split in a training set of 60,000 examples and a test set of 10,000 examples. Each image is a grayscale image associated, as the MNIST dataset, with a label from 10 classes, from 0 to 9. Each example represents a different item of clothing. As for the MNIST dataset, the training and the following test phase were conducted on each class, one at a time.

\textbf{CIFAR-10:}
A dataset of tiny images of various subjects, CIFAR-10 consists of 60000  $32 \times 32$ color images in 10 classes, with 6000 images per class. There are 50000 training images and 10000 test images.

\textbf{KDD:} KDD dataset \cite{Lichman2013} consists in a collection of network intrusion detection data, this is the data set used for The Third International Knowledge Discovery and Data Mining Tools Competition held in conjunction with KDD-99 The Fifth International Conference on Knowledge Discovery and Data Mining. The competition task was to build a network intrusion detector, a predictive model capable of distinguishing between ``bad'' connections, called intrusions or attacks, and ``good'' normal connections. All the examples are string tuples or numbers. Given the considerable amount of data inside the data set, we used as in \cite{Zenati2018}, the KDDCUP99 10 percent version of the data set. Since the anomalies are in a higher percentage than normal data, the normal ones have been considered as anomalies.

\subsubsection{Methodology}
We follow the same methodology employed by Zenati et al. in the official code accompanying \citet{Zenati2018}.

We start by getting all the dataset together (train and test split). Starting from this one big pool of examples, we choose one class as an anomaly and, after shuffling the dataset, we then create a training set by using 80\% of the whole data while the remaining 20\% is used for the testing set, this process is repeated for all the classes in the dataset. Each model is trained accordingly to its original implementation on the training set and is tested on the whole dataset made up of both regular and anomalous data.

\subsection{Results}
\label{sec:results}
All tests have been made measuring the area under precision and recall curve (AUPRC). For a complete understanding of all the results, please refer to \cref{sec:appendix}. Our fully static-Tensorflow implementation (i.e., only Tensorflow framework has been used, we relied upon neither Keras nor eager execution) produced different results from the one depicted in the original \citet{Donahue2016}, \citet{Schlegl2017}, and \citet{Akcay2018} papers. The tests have been made by training the GAN networks with different hyper-parameters configurations in order to test a broader range of models configurations. Moreover, following the GANomaly approach, we have added an evaluation process to make model selection and thus select the best model during the training phase. The model selection has been made for BiGAN and GANomaly, since doing it for AnoGAN would be unfeasible, due to the $\Gamma$ research steps required for every sample at test-time.
We intentionally left out the performance evaluation of the AnoGAN model. Due to the inner workings of the architecture that should have required a very long time to be tested because of the necessity to find, every time and for every image taken into consideration, the best latent representation, meaning that for every image we would need 500 training step (500 it is an empirical value found in the original paper \cite{Schlegl2017}).
Following, the training and test methods are described. A summary of the results is present in \cref{img:BiGAN_GANomaly_resutls} for the MNIST and Fashion-MNIST dataset, and in \cref{tab:kdd_results} for the KDD results.

\begin{table}
	\centering
	\rowcolors[]{1}{white}{gray!25}
	\begin{tabular}{@{}lcccccc@{}}  \toprule
		       & \multicolumn{3}{c}{Train} &            & \multicolumn{2}{c}{Test}                              \\ \cmidrule(lr){2-4} \cmidrule(l){6-7}
		\rowcolor{white}
		       & BCE                       & FM         & Residual                 &  & BCE        & FM         \\ \midrule
		Case 1 & \checkmark                &            & \checkmark               &  & \checkmark & \checkmark \\

		Case 2 &                           & \checkmark & \checkmark               &  & \checkmark & \checkmark \\

		Case 3 & \checkmark                &            &                          &  & \checkmark & \checkmark \\

		Case 4 &                           & \checkmark &                          &  & \checkmark & \checkmark \\
		\bottomrule
	\end{tabular}
	\caption{BiGAN - Training and testing configuration combinations.}
	\label{tab:BiGAN_config_comb}
\end{table}

\textbf{BiGAN/EGBAD:}
With BiGAN/EGBAD \cite{Zenati2018} architecture we have executed the highest number of training and testing configurations, this means we have performed, for every label the combinations depicted in \cref{tab:BiGAN_config_comb}.
To better understand what the table shows, taking the first row (Case 1) as an example, we are describing the case where, during the training phase, a Binary Cross Entropy loss (BCE) in combination with a Residual loss has been used and, during the testing phase, we computed the anomaly score twice, using the feature matching (FM) and the BCE. All the other cases should be clear.
In particular, during the training phase, the choice between using BCE or FM influences only the generator loss. The residual loss that it is intended as the difference between the original image and the image reconstructed by the generator starting from a latent representation (please refer to \cref{residual_loss}) when used has been added as an additional term to the BCE loss of the encoder.
In \cref{img:BiGAN_GANomaly_resutls} is possible to see the results on the MNIST and Fashion-MNIST datasets. In \cref{sec:appendix} it is possible to see the complete results. The images used are the original $28\times 28$ images. Any additional result from multiple different tests performed with different random seeds has been omitted for the sake of clarity.

\textbf{GANomaly:}
GANomaly training and testing have been done similarly to the previously described BiGAN/EGBAD architecture. The whole procedure is leaner here, with only the combinations shown in \cref{tab:GANomaly_config_comb}.

\begin{table}
	\centering
	\rowcolors[]{1}{white}{gray!25}
	\begin{tabular}{@{}lcccccc@{}}  \toprule
		       & \multicolumn{2}{c}{Train} &            & \multicolumn{2}{c}{Test}                           \\ \cmidrule(lr){2-3} \cmidrule(l){5-6}
		\rowcolor{white}
		       & BCE                       & FM         &                          & BCE        & FM         \\ \midrule

		Case 1 & \checkmark                &            &                          & \checkmark & \checkmark \\
		Case 2 &                           & \checkmark &                          & \checkmark & \checkmark \\
		\bottomrule
	\end{tabular}
	\caption{GANomaly - Training and testing configuration combinations.}
	\label{tab:GANomaly_config_comb}
\end{table}

We added a step of model selection during the training phase in order to always save the very best model. For this architecture, the testing phase has been done using an anomaly score equal to the squared difference between the latent representations of the image encoded first with autoencoder part of the network and, after being reconstructed, encoded again with the encoder. To briefly review the working of GANomaly, see \cref{fig:ganomaly_pipeline}. The images used are the original ones resized to $32\times 32$. Refer to the plot on the right of \cref{img:BiGAN_GANomaly_resutls} to a  brief summary of the results on MNIST and Fashion-MNIST datasets and on \cref{tab:kdd_results} for the results on KDD dataset.

\textbf{AnoGAN:}
As previously stated, the results of AnoGAN have not been reproduced due to the onerous computational requirements causing the train and test phases to be intensively time-consuming. The reader could refer to the original paper \cite{Schlegl2017} to an overview of the results.

\begin{table}
	\centering
	\rowcolors[]{1}{white}{gray!25}
	\begin{tabular}{@{}lcccccc@{}}  \toprule
		            & \multicolumn{3}{c}{KDD}                                         \\ \cmidrule(lr){2-4}
		\rowcolor{white}
		            & Precision               & Recall            & F1-Score          \\ \midrule

		BiGAN/EGBAD & \textbf{0.941174}       & \textbf{0.956155} & \textbf{0.948605} \\
		GANomaly    & 0.830256                & 0.841112          & 0.835648          \\
		\bottomrule
	\end{tabular}
	\caption{The performances of BiGAN/EGBAD and GANomaly models on the KDD dataset.}
	\label{tab:kdd_results}
\end{table}

\section{Conclusions}
\label{sec:conclusions}
We implemented and evaluated the state-of-the-art algorithms for anomaly detection based on GANs. In this work, we unified the three major works under the same framework (Tensorflow) and within the same code-base. The analysis and implementation of the algorithms allowed us to verify the effectiveness of the GANs-based approach to the anomaly detection problem and, at the same time, highlight the differences between the original papers and the publicly available code. Besides, to test the feasibility of the architectures mentioned above, this work intends to provide a modular and ready-to-go solution for everyone desiring an anomaly detection toolkit working out of the box.
\bibliography{bib}
\bibliographystyle{icml2019}
\clearpage
\onecolumn
\appendix
\section{Additional Results}
\label{sec:appendix}

In this section, we present more detailed results regarding the different combinations of training-testing pipelines of the BiGAN/EGBAD and GANomaly architectures on MNIST and Fashion-MNIST; results of the CIFAR10 dataset, trained with GANomaly are presented too.
Moreover, the results of the addition of a function (namely, a leaky relu activation function in the first layer of the encoder) that was missing from inside the EGBAD published code architecture will be introduced. Surprisingly, the tests have led to the result that without that activation function the network performs better.

\subsection{Version 1}
\label{sec:version_1}
Before getting the results shown in \cref{sec:version_2} we have employed a different test pipeline with a different architecture of the BiGAN network. The main difference was mainly related to:
\begin{itemize}
	\itemsep-3em
	      \vspace{-1cm}
	\item \textbf{Testing phase} The testing phase is applied twice (for both BiGAN/EGBAD and GANomaly architectures). Once during the model selection during training, and once after the training was finished as a stand alone phase.
	\item \textbf{Activation function} A Leaky Relu function, erroneously missing from the original EGBAD \cite{Zenati2018} work has been here inserted after the first convolutional layer of the encoder model.
\end{itemize}
\vspace{-1cm}
A brief visual understanding of the configuration cases for the BiGAN/EGBAD architecture it is shown in \cref{tab:BiGAN_config_comb_old} below. The configuration is similar to the one shown in \cref{tab:BiGAN_config_comb} with the difference that, here, the testing phase has been done separately. The same is true for the GANomaly implemented architecture, shown in \cref{tab:GANomaly_config_comb_old}.

\begin{table}[!htb]
	\centering
	\rowcolors[]{1}{white}{gray!25}

	\begin{tabular}{@{}lcccccc@{}}  \toprule
		       & \multicolumn{3}{c}{Train} &            & \multicolumn{2}{c}{Test}                              \\ \cmidrule(lr){2-4} \cmidrule(l){6-7}
		\rowcolor{white}
		       & BCE                       & FM         & Residual                 &  & BCE        & FM         \\ \midrule
		Case 1 & \checkmark                &            & \checkmark               &  & \checkmark &            \\
		Case 2 & \checkmark                &            & \checkmark               &  &            & \checkmark \\
		Case 3 &                           & \checkmark & \checkmark               &  & \checkmark &            \\
		Case 4 &                           & \checkmark & \checkmark               &  &            & \checkmark \\
		Case 5 & \checkmark                &            &                          &  & \checkmark &            \\
		Case 6 & \checkmark                &            &                          &  &            & \checkmark \\
		Case 7 &                           & \checkmark &                          &  & \checkmark &            \\
		Case 8 &                           & \checkmark &                          &  &            & \checkmark \\
		\bottomrule
	\end{tabular}
	\caption{BiGAN - Training and testing configuration combinations.}
	\label{tab:BiGAN_config_comb_old}
\end{table}
\vspace{-3cm}
\begin{table}
	\centering
	\rowcolors[]{1}{white}{gray!25}
	\begin{tabular}{@{}lcccccc@{}}  \toprule
		       & \multicolumn{2}{c}{Train} &            & \multicolumn{2}{c}{Test}                           \\ \cmidrule(lr){2-3} \cmidrule(l){5-6}
		\rowcolor{white}
		       & BCE                       & FM         &                          & BCE        & FM         \\ \midrule

		Case 5 & \checkmark                &            &                          & \checkmark &            \\
		Case 6 & \checkmark                &            &                          &            & \checkmark \\
		Case 7 &                           & \checkmark &                          & \checkmark &            \\
		Case 8 &                           & \checkmark &                          &            & \checkmark \\
		\bottomrule
	\end{tabular}
	\caption{GANomaly - Training and testing configuration combinations.}
	\label{tab:GANomaly_config_comb_old}
\end{table}

\pagebreak
\subsubsection{MNIST}
\begin{figure}[!htb]
\centering
{\centering \textbf{GANomaly test results on MNIST dataset}\par\medskip}
\begin{tikzpicture}[scale=1,line width=1pt, mark size=3pt]
\begin{axis}[
    xlabel={(a) Anomalous classes},
    ylabel={AUPRC},
    xmin=-0.5, xmax=9.5,
    ymin=0, ymax=1.2,
    xtick={0, 1, 2, 3, 4, 5, 6, 7, 8, 9},
    ytick={0.2, 0.4, 0.6, 0.8, 1},
    legend pos=north west,
    legend style={font=\fontsize{6}{5}\selectfont},
    ymajorgrids=true,
    grid style=dashed,
]
    \addplot[ 
    only marks,
    color=red,
    mark=triangle,
    error bars/.cd, y dir=both,y explicit
    ]
    coordinates {
    (0,0.674329)
	(1,0.248704)
	(2,0.746328)
	(3,0.586607)
	(4,0.540687)
	(5,0.579946)
	(6,0.699154)
	(7,0.376821)
	(8,0.710517)
	(9,0.400182)
    };
    
    \addplot[ 
    only marks,
    color=green,
    mark=square,
    error bars/.cd, y dir=both,y explicit
    ]
    coordinates {
    (0,0.667248)
	(1,0.248307)
	(2,0.732796)
	(3,0.568799)
	(4,0.568901)
	(5,0.603012)
	(6,0.669514)
	(7,0.401656)
	(8,0.702614)
	(9,0.398312)
    };    
    \legend{MNIST bce, MNIST fm}
\end{axis}
\end{tikzpicture}
\caption{GANomaly results on MNIST. All tests are here performed through a bce metric and measured by the area under the precision-recall curve (AURPC). Training is done with a bce loss and a fm loss.}
\label{fig:ganomaly_mnist_results_v1}
\end{figure}
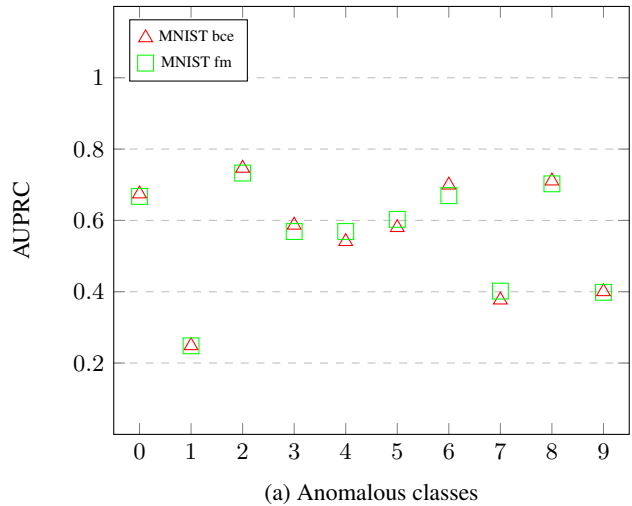

Here, we present the results on the MNIST dataset. We show in \cref{img:BiGAN_mnist_v1} the results of the BiGA/EGBAD architecture and in \cref{fig:ganomaly_mnist_results_v1} the results of the GANomaly architecture.
In the following, MNIST results are depicted.
\begin{figure}[!htb]
\centering
{\centering \textbf{MNIST Tests (w and w/o residual loss)}\par\medskip}
\begin{tikzpicture}[scale=1,line width=1pt, mark size=3pt]
\begin{axis}[
    xlabel={(a) Anomalous classes},
    ylabel={AUPRC},
    xmin=-0.5, xmax=9.5,
    ymin=0, ymax=0.85,
    xtick={0, 1, 2, 3, 4, 5, 6, 7, 8, 9},
    ytick={0.2, 0.4, 0.6, 0.8, 1},
    legend pos=north east,
    legend style={font=\fontsize{6}{5}\selectfont},
    ymajorgrids=true,
    grid style=dashed,
]
    \addplot[ 
    only marks,
    color=red,
    mark=triangle,
    error bars/.cd, y dir=both,y explicit
    ]
    coordinates {
    (0,0.588973)
	(1,0.945718)
	(2,0.673012)
	(3,0.328175)
	(4,0.549944)
	(5,0.373951)
	(6,0.344472)
	(7,0.159381)
	(8,0.410359)
	(9,0.204547)
    };
    
    \addplot[ 
    only marks,
    color=green,
    mark=square,
    error bars/.cd, y dir=both,y explicit
    ]
    coordinates {
    (0,0.576642)
	(1,0.946169)
	(2,0.677170)
	(3,0.330465)
	(4,0.557133)
	(5,0.366146)
	(6,0.335983)
	(7,0.169242)
	(8,0.402505)
	(9,0.202300)
    };
    
    \addplot[ 
    only marks,
    color=blue,
    mark=star,
    error bars/.cd, y dir=both,y explicit
    ]
    coordinates {
    (0,0.532921)
	(1,0.945711)
	(2,0.631290)
	(3,0.291819)
	(4,0.459531)
	(5,0.388838)
	(6,0.378829)
	(7,0.169274)
	(8,0.479645)
	(9,0.183675)
    };
    
    \addplot[ 
    only marks,
    color=yellow,
    opacity=0.85,
    mark=*,
    error bars/.cd, y dir=both,y explicit
    ]
    coordinates {
    (0,0.491745)
	(1,0.946415)
	(2,0.629301)
	(3,0.280620)
	(4,0.482244)
	(5,0.385905)
	(6,0.381410)
	(7,0.177437)
	(8,0.485829)
	(9,0.189037)
    };
    
    \legend{brtb, brtf, frtb, frtf}
\end{axis}
\end{tikzpicture}
\begin{tikzpicture}[scale=1,line width=1pt, mark size=3pt]
\begin{axis}[
    xlabel={(b) Anomalous classes},
    ylabel={AUPRC},
    xmin=-0.5, xmax=9.5,
    ymin=0, ymax=0.85,
    xtick={0, 1, 2, 3, 4, 5, 6, 7, 8, 9},
    ytick={0.2, 0.4, 0.6, 0.8, 1},
    legend pos=north east,
    legend style={font=\fontsize{6}{5}\selectfont},
    ymajorgrids=true,
    grid style=dashed,
]
    \addplot[ 
    only marks,
    color=red,
    mark=triangle,
    error bars/.cd, y dir=both,y explicit
    ]
    coordinates {
    (0,0.460489)
	(1,0.945778)
	(2,0.488333)
	(3,0.252856)
	(4,0.233408)
	(5,0.189892)
	(6,0.386621)
	(7,0.370795)
	(8,0.337458)
	(9,0.171530)

    };
    
    \addplot[ 
    only marks,
    color=green,
    mark=square,
    error bars/.cd, y dir=both,y explicit
    ]
    coordinates {
    (0,0.473931)
	(1,0.945785)
	(2,0.483630)
	(3,0.243344)
	(4,0.235263)
	(5,0.184278)
	(6,0.381939)
	(7,0.390189)
	(8,0.333909)
	(9,0.173908)
    };
    
    \addplot[ 
    only marks,
    color=blue,
    mark=star,
    error bars/.cd, y dir=both,y explicit
    ]
    coordinates {
    (0,0.576616)
	(1,0.945824)
	(2,0.398558)
	(3,0.269676)
	(4,0.214738)
	(5,0.206684)
	(6,0.384334)
	(7,0.201369)
	(8,0.274019)
	(9,0.164905)
    };
    
    \addplot[ 
    only marks,
    color=yellow,
    opacity=0.85,
    mark=*,
    error bars/.cd, y dir=both,y explicit
    ]
    coordinates {
    (0,0.549113)
	(1,0.946829)
	(2,0.394120)
	(3,0.259436)
	(4,0.222208)
	(5,0.200468)
	(6,0.378746)
	(7,0.208283)
	(8,0.266873)
	(9,0.170369)
    };
    
    \legend{btb, btf, ftb, ftf}
\end{axis}
\end{tikzpicture}
\caption{Performance of version 1 of BiGAN/EGBAD on MNIST measured by the area under the precision-recall curve (AURPC). Best viewed in color. Image (a) depicts the results using the residual loss, image (b) it is without the residual loss. "b(r)tb": bce (+ residual loss) trained and bce tested, "b(r)tf": bce (+ residual loss) trained and fm tested, "f(r)tb": fm (+ residual loss) trained and bce tested, "f(r)tf": fm (+ residual loss) trained and fm tested}
\label{img:BiGAN_mnist_v1}
\end{figure}
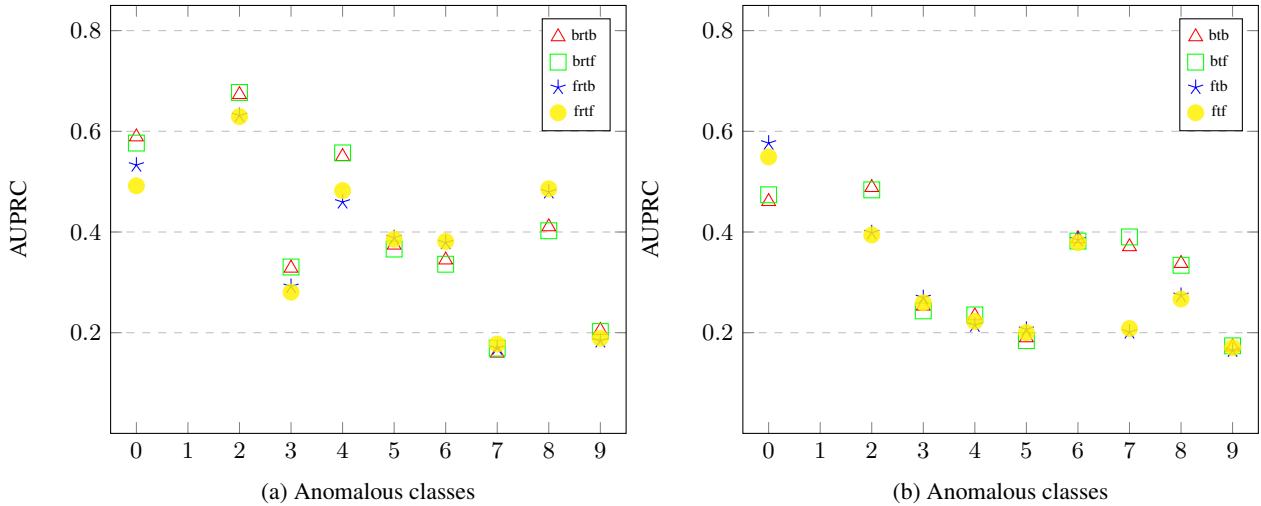

\clearpage

\subsubsection{Fashion-MNIST}
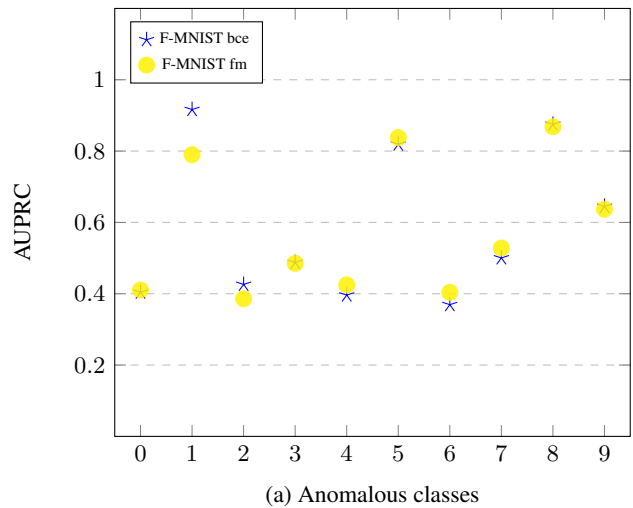
\begin{figure}[!htb]
\centering
{\centering \textbf{GANomaly test results on Fashion-MNIST dataset}\par\medskip}
\begin{tikzpicture}[scale=1,line width=1pt, mark size=3pt]
\begin{axis}[
    xlabel={(a) Anomalous classes},
    ylabel={AUPRC},
    xmin=-0.5, xmax=9.5,
    ymin=0, ymax=1.2,
    xtick={0, 1, 2, 3, 4, 5, 6, 7, 8, 9},
    ytick={0.2, 0.4, 0.6, 0.8, 1},
    legend pos=north west,
    legend style={font=\fontsize{6}{5}\selectfont},
    ymajorgrids=true,
    grid style=dashed,
]   
    \addplot[ 
    only marks,
    color=blue,
    mark=star,
    error bars/.cd, y dir=both,y explicit
    ]
    coordinates {
    (0,0.404195)
	(1,0.916438)
	(2,0.426077)
	(3,0.487566)
	(4,0.396193)
	(5,0.819338)
	(6,0.369295)
	(7,0.500377)
	(8,0.875034)
	(9,0.644799)
    };
    
    \addplot[ 
    only marks,
    color=yellow,
    opacity=0.85,
    mark=*,
    error bars/.cd, y dir=both,y explicit
    ]
    coordinates {
    (0,0.410540)
	(1,0.789967)
	(2,0.385870)
	(3,0.485164)
	(4,0.424754)
	(5,0.837917)
	(6,0.403830)
	(7,0.528114)
	(8,0.868231)
	(9,0.637120)
    };
    \legend{F-MNIST bce, F-MNIST fm}
\end{axis}
\end{tikzpicture}
\caption{GANomaly results on Fashion-MNIST. All tests are here performed with a bce metric and using the area under the precision-recall curve (AURPC). Training is done with a bce loss and a fm loss.}
\label{fig:ganomaly_fashion_mnist_results_v1}
\end{figure}

Here we present the results obtained on the Fashion-MNIST dataset. In \cref{img:BiGAN_fashion_mnist_v1} we show the results of BiGAN/EGBAD architectures and in \cref{fig:ganomaly_fashion_mnist_results_v1} the results of the GANomaly architecture.
\begin{figure*}[!htb]
\centering
{\centering \textbf{Fashion-MNIST Tests (w and w/o residual loss)}\par\medskip}
\begin{tikzpicture}[scale=1,line width=1pt, mark size=3pt]
\begin{axis}[
    xlabel={(a) Anomalous classes},
    ylabel={AUPRC},
    xmin=-0.5, xmax=9.5,
    ymin=0, ymax=0.85,
    xtick={0, 1, 2, 3, 4, 5, 6, 7, 8, 9},
    ytick={0.2, 0.4, 0.6, 0.8, 1},
    legend pos=north west,
    legend style={font=\fontsize{6}{5}\selectfont},
    ymajorgrids=true,
    grid style=dashed,
]
    \addplot[ 
    only marks,
    color=red,
    mark=triangle,
    error bars/.cd, y dir=both,y explicit
    ]
    coordinates {
    (0,0.329793)
	(1,0.395117)
	(2,0.326721)
	(3,0.407941)
	(4,0.315398)
	(5,0.283340)
	(6,0.231656)
	(7,0.239397)
	(8,0.700767)
	(9,0.539585)
    };
    
    \addplot[ 
    only marks,
    color=green,
    mark=square,
    error bars/.cd, y dir=both,y explicit
    ]
    coordinates {
    (0,0.326310)
	(1,0.422022)
	(2,0.331581)
	(3,0.406020)
	(4,0.308274)
	(5,0.287235)
	(6,0.235714)
	(7,0.250416)
	(8,0.684940)
	(9,0.537952)
    };
    
    \addplot[ 
    only marks,
    color=blue,
    mark=star,
    error bars/.cd, y dir=both,y explicit
    ]
    coordinates {
    (0,0.304585)
	(1,0.465419)
	(2,0.293363)
	(3,0.344648)
	(4,0.311827)
	(5,0.279147)
	(6,0.226567)
	(7,0.245187)
	(8,0.700235)
	(9,0.447232)
    };
    
    \addplot[ 
    only marks,
    color=yellow,
    opacity=0.85,
    mark=*,
    error bars/.cd, y dir=both,y explicit
    ]
    coordinates {
    (0,0.310653)
	(1,0.505422)
	(2,0.303262)
	(3,0.327715)
	(4,0.303162)
	(5,0.290173)
	(6,0.227067)
	(7,0.247125)
	(8,0.695714)
	(9,0.439202)
    };
    
    \legend{brtb, brtf, frtb, frtf}
\end{axis}
\end{tikzpicture}
\begin{tikzpicture}[scale=1,line width=1pt, mark size=3pt]
\begin{axis}[
    xlabel={(b) Anomalous classes},
    ylabel={AUPRC},
    xmin=-0.5, xmax=9.5,
    ymin=0, ymax=0.85,
    xtick={0, 1, 2, 3, 4, 5, 6, 7, 8, 9},
    ytick={0.2, 0.4, 0.6, 0.8, 1},
    legend pos=north west,
    legend style={font=\fontsize{6}{5}\selectfont},
    ymajorgrids=true,
    grid style=dashed,
]
    \addplot[ 
    only marks,
    color=red,
    mark=triangle,
    error bars/.cd, y dir=both,y explicit
    ]
    coordinates {
    (0,0.334616)
	(1,0.345568)
	(2,0.266158)
	(3,0.309683)
	(4,0.278540)
	(5,0.354789)
	(6,0.193593)
	(7,0.377128)
	(8,0.526488)
	(9,0.529023)
    };
    
    \addplot[ 
    only marks,
    color=green,
    mark=square,
    error bars/.cd, y dir=both,y explicit
    ]
    coordinates {
    (0,0.332080)
	(1,0.357153)
	(2,0.269760)
	(3,0.305771)
	(4,0.280468)
	(5,0.279281)
	(6,0.193988)
	(7,0.365489)
	(8,0.532075)
	(9,0.523796)
    };
    
    \addplot[ 
    only marks,
    color=blue,
    mark=star,
    error bars/.cd, y dir=both,y explicit
    ]
    coordinates {
    (0,0.136560)
	(1,0.368872)
	(2,0.347168)
	(3,0.190338)
	(4,0.240090)
	(5,0.403063)
	(6,0.200473)
	(7,0.512209)
	(8,0.497529)
	(9,0.542557)
    };
    
    \addplot[ 
    only marks,
    color=yellow,
    opacity=0.85,
    mark=*,
    error bars/.cd, y dir=both,y explicit
    ]
    coordinates {
    (0,0.135812)
	(1,0.386817)
	(2,0.348158)
	(3,0.193076)
	(4,0.238491)
	(5,0.402083)
	(6,0.204106)
	(7,0.516749)
	(8,0.495293)
	(9,0.527323)
    };
    
    \legend{btb, btf, ftb, ftf}
\end{axis}
\end{tikzpicture}
\caption{Performance of version 1 of BiGAN/EGBAD on Fashion-MNIST measured by the area under the precision-recall curve (AURPC). Best viewed in color. Image (a) depicts the results using the residual loss, image (b) it is without the residual loss. "b(r)tb": bce (+ residual loss) trained and bce tested, "b(r)tf": bce (+ residual loss) trained and fm tested, "f(r)tb": fm (+ residual loss) trained and bce tested, "f(r)tf": fm (+ residual loss) trained and fm tested}
\label{img:BiGAN_fashion_mnist_v1}
\end{figure*}

\clearpage

\subsubsection{CIFAR10}
\begin{figure}[!htb]
\centering
{\centering \textbf{GANomaly test results on Cifar10 dataset}\par\medskip}
\begin{tikzpicture}[scale=1,line width=1pt, mark size=3pt]
\begin{axis}[
    xlabel={(a) Anomalous classes},
    ylabel={AUPRC},
    xmin=-0.5, xmax=9.5,
    ymin=0, ymax=0.85,
    xtick={0, 1, 2, 3, 4, 5, 6, 7, 8, 9},
    ytick={0.2, 0.4, 0.6, 0.8, 1},
    legend pos=north west,
    legend style={font=\fontsize{6}{5}\selectfont},
    ymajorgrids=true,
    grid style=dashed,
]   
    \addplot[ 
    only marks,
    color=darkgray,
    mark=+,
    error bars/.cd, y dir=both,y explicit
    ]
    coordinates {
	(0,0.426117)
	(1,0.509281)
	(2,0.308877)
	(3,0.423059)
	(4,0.273186)
	(5,0.418272)
	(6,0.323785)
	(7,0.378428)
	(8,0.399429)
	(9,0.508734)
    };
    
    \addplot[ 
    only marks,
    color=olive,
    mark=pentagon,
    error bars/.cd, y dir=both,y explicit
    ]
    coordinates {
	(0,0.419792)
	(1,0.499168)
	(2,0.299644)
	(3,0.412572)
	(4,0.269214)
	(5,0.404921)
	(6,0.296456)
	(7,0.389358)
	(8,0.394945)
	(9,0.513593)
    };
    
    \legend{MNIST bce, MNIST fm, F-MNIST bce, F-MNIST fm, Cifar10 bce, Cifar10 fm}
\end{axis}
\end{tikzpicture}
\caption{GANomaly results on CIFAR10. All tests are here performed through a bce metric. Training is done with a bce loss and a fm loss.}
\label{fig:ganomaly_cifar10_results_v1}
\end{figure}
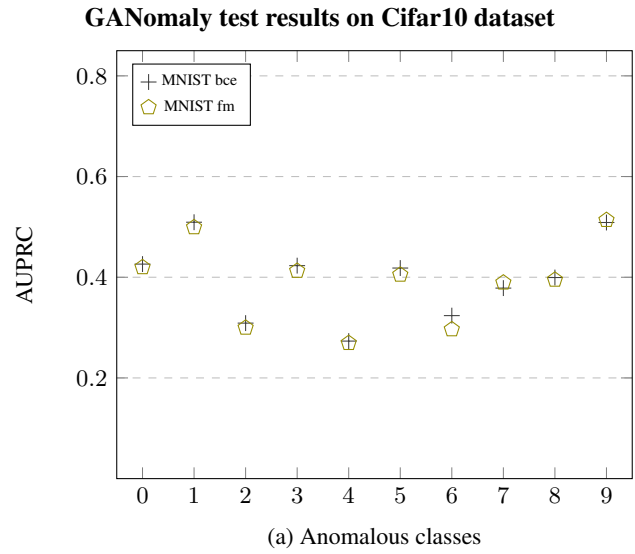

In the following \cref{fig:ganomaly_cifar10_results_v1} we present the results on the CIFAR10 dataset.
\clearpage

\subsection{Version 2}
\label{sec:version_2}
In \cref{sec:version_2} we get back to the original version of the EGBAD \cite{Zenati2018} and find that the results are much better, even if there is no activation function in the first convolutional layer. The main differences from \cref{sec:version_1} are the following:
\begin{itemize}
	\item \textbf{Testing phase} The testing phase is applied only once during the model selection during training.
	\item \textbf{Activation function} The Leaky Relu function has here been left out as for the original EGBAD paper\cite{Zenati2018}.
\end{itemize}

The training and testing pipelines employed have been already described. Please refer to section \ref{sec:bigan} and section \ref{sec:ganomaly} for more details. Every following section will introduce the results for a specific dataset.

\subsubsection{MNIST}
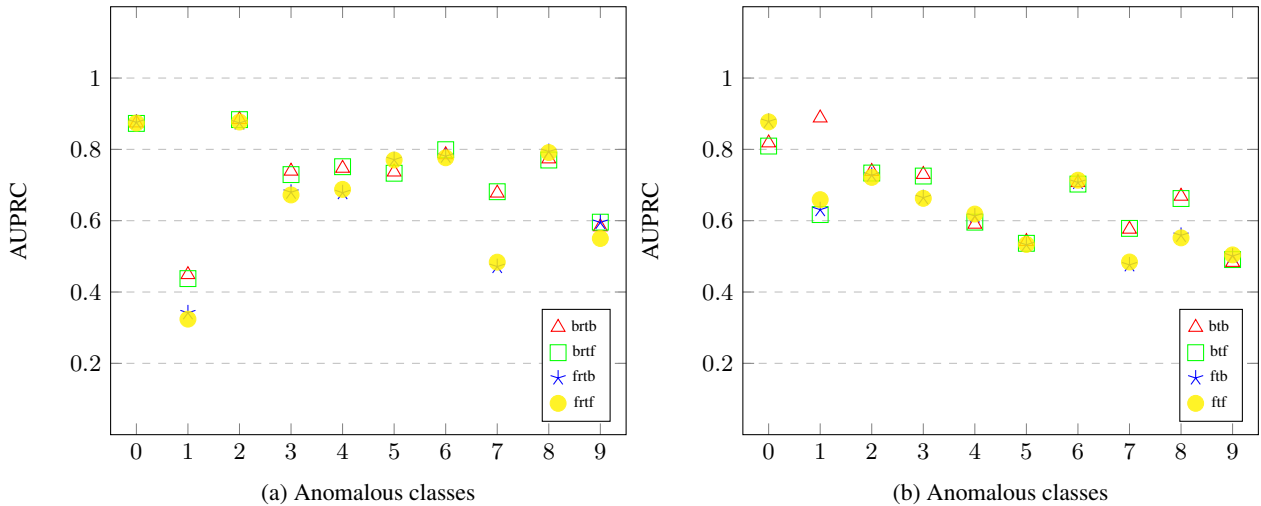
\begin{figure}[!htb]
\centering
{\centering \textbf{MNIST Tests (w and w/o residual loss)}\par\medskip}
\begin{tikzpicture}[scale=1,line width=1pt, mark size=3pt]
\begin{axis}[
    xlabel={(a) Anomalous classes},
    ylabel={AUPRC},
    xmin=-0.5, xmax=9.5,
    ymin=0, ymax=1.2,
    xtick={0, 1, 2, 3, 4, 5, 6, 7, 8, 9},
    ytick={0.2, 0.4, 0.6, 0.8, 1},
    legend pos=south east,
    legend style={font=\fontsize{6}{5}\selectfont},
    ymajorgrids=true,
    grid style=dashed,
]
    \addplot[ 
    only marks,
    color=red,
    mark=triangle,
    error bars/.cd, y dir=both,y explicit
    ]
    coordinates {
    (0,0.874195)
	(1,0.448800)
	(2,0.883708)
	(3,0.738898)
	(4,0.747029)
	(5,0.736488)
	(6,0.784979)
	(7,0.677388)
	(8,0.773257)
	(9,0.582801)
    };
    
    \addplot[ 
    only marks,
    color=green,
    mark=square,
    error bars/.cd, y dir=both,y explicit
    ]
    coordinates {
    (0,0.872826)
	(1,0.437619)
	(2,0.883426)
	(3,0.729275)
	(4,0.751384)
	(5,0.733082)
	(6,0.799101)
	(7,0.681276)
	(8,0.770090)
	(9,0.595880)
    };
    
    \addplot[ 
    only marks,
    color=blue,
    mark=star,
    error bars/.cd, y dir=both,y explicit
    ]
    coordinates {
    (0,0.875797)
	(1,0.341691)
	(2,0.874077)
	(3,0.679769)
	(4,0.678516)
	(5,0.771094)
	(6,0.780822)
	(7,0.471408)
	(8,0.793888)
	(9,0.595137)
    };
    
    \addplot[ 
    only marks,
    color=yellow,
    opacity=0.85,
    mark=*,
    error bars/.cd, y dir=both,y explicit
    ]
    coordinates {
    (0,0.873892)
	(1,0.323952)
	(2,0.876721)
	(3,0.672212)
	(4,0.687852)
	(5,0.770176)
	(6,0.776446)
	(7,0.483706)
	(8,0.791080)
	(9,0.550133)
    };
    
    \legend{brtb, brtf, frtb, frtf}
\end{axis}
\end{tikzpicture}
\begin{tikzpicture}[scale=1,line width=1pt, mark size=3pt]
\begin{axis}[
    xlabel={(b) Anomalous classes},
    ylabel={AUPRC},
    xmin=-0.5, xmax=9.5,
    ymin=0, ymax=1.2,
    xtick={0, 1, 2, 3, 4, 5, 6, 7, 8, 9},
    ytick={0.2, 0.4, 0.6, 0.8, 1},
    legend pos=south east,
    legend style={font=\fontsize{6}{5}\selectfont},
    ymajorgrids=true,
    grid style=dashed,
]
    \addplot[ 
    only marks,
    color=red,
    mark=triangle,
    error bars/.cd, y dir=both,y explicit
    ]
    coordinates {
    (0,0.818171)
	(1,0.888087)
	(2,0.738119)
	(3,0.729733)
	(4,0.590095)
	(5,0.541676)
	(6,0.706885)
	(7,0.576134)
	(8,0.668595)
	(9,0.483049)
    };
    
    \addplot[ 
    only marks,
    color=green,
    mark=square,
    error bars/.cd, y dir=both,y explicit
    ]
    coordinates {
    (0,0.809255)
	(1,0.617134)
	(2,0.733652)
	(3,0.725185)
	(4,0.595918)
	(5,0.536608)
	(6,0.702646)
	(7,0.578404)
	(8,0.662140)
	(9,0.490905)
    };
    
    \addplot[ 
    only marks,
    color=blue,
    mark=star,
    error bars/.cd, y dir=both,y explicit
    ]
    coordinates {
    (0,0.878393)
	(1,0.630889)
	(2,0.727515)
	(3,0.664605)
	(4,0.613168)
	(5,0.532516)
	(6,0.707204)
	(7,0.476107)
	(8,0.558868)
	(9,0.500981)
    };
    
    \addplot[ 
    only marks,
    color=yellow,
    opacity=0.85,
    mark=*,
    error bars/.cd, y dir=both,y explicit
    ]
    coordinates {
    (0,0.877100)
	(1,0.658987)
	(2,0.721466)
	(3,0.662677)
	(4,0.618580)
	(5,0.533485)
	(6,0.713523)
	(7,0.484160)
	(8,0.551857)
	(9,0.504099)
    };
    
    \legend{btb, btf, ftb, ftf}
\end{axis}
\end{tikzpicture}
\caption{Performance of version 2 of BiGAN/EGBAD on MNIST measured by the area under the precision-recall curve (AUPRC). Best viewed in color. Image (a) depicts the results using the residual loss, image (b) it is without the residual loss. "b(r)tb": bce (+ residual loss) trained and bce tested, "b(r)tf": bce (+ residual loss) trained and fm tested, "f(r)tb": fm (+ residual loss) trained and bce tested, "f(r)tf": fm (+ residual loss) trained and fm tested}
\label{img:BiGAN_mnist_v2}
\end{figure}

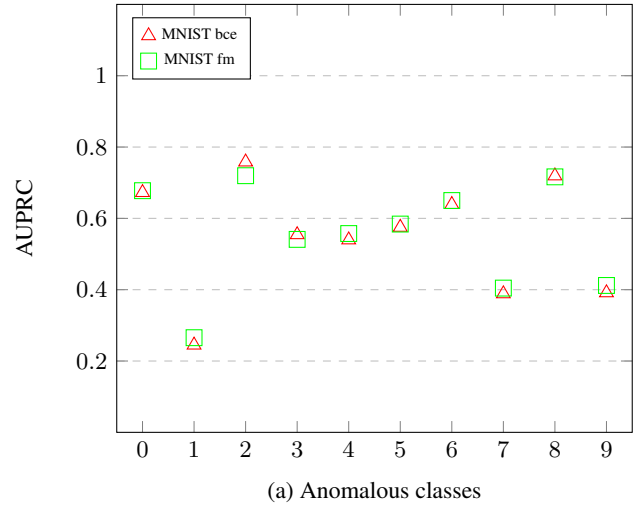
\begin{figure}[!htb]
\centering
{\centering \textbf{GANomaly test results on MNIST dataset}\par\medskip}
\begin{tikzpicture}[scale=1,line width=1pt, mark size=3pt]
\begin{axis}[
    xlabel={(a) Anomalous classes},
    ylabel={AUPRC},
    xmin=-0.5, xmax=9.5,
    ymin=0, ymax=1.2,
    xtick={0, 1, 2, 3, 4, 5, 6, 7, 8, 9},
    ytick={0.2, 0.4, 0.6, 0.8, 1},
    legend pos=north west,
    legend style={font=\fontsize{6}{5}\selectfont},
    ymajorgrids=true,
    grid style=dashed,
]
    \addplot[ 
    only marks,
    color=red,
    mark=triangle,
    error bars/.cd, y dir=both,y explicit
    ]
    coordinates {
    (0,0.672449)
	(1,0.244729)
	(2,0.758134)
	(3,0.554249)
	(4,0.539690)
	(5,0.574999)
	(6,0.638957)
	(7,0.388524)
	(8,0.719157)
	(9,0.391370)
    };
    
    \addplot[ 
    only marks,
    color=green,
    mark=square,
    error bars/.cd, y dir=both,y explicit
    ]
    coordinates {
    (0,0.677662)
	(1,0.265432)
	(2,0.719683)
	(3,0.541056)
	(4,0.557419)
	(5,0.584021)
	(6,0.649983)
	(7,0.404483)
	(8,0.716441)
	(9,0.411862)
    };    
    \legend{MNIST bce, MNIST fm}
\end{axis}
\end{tikzpicture}
\caption{GANomaly results on MNIST. All tests are here performed through a bce metric and using the area under the precision-recall curve (AURPC). Training is done with a bce loss and a fm loss.}
\label{fig:ganomaly_mnist_results_v2}
\end{figure}

Here we present the results for the MNIST dataset. In \cref{img:BiGAN_mnist_v2} we show the results of BiGAN/EGBAD. In \cref{fig:ganomaly_mnist_results_v2} we show the results of GANomaly architecture.
\clearpage

\clearpage

\subsubsection{Fashion-MNIST}
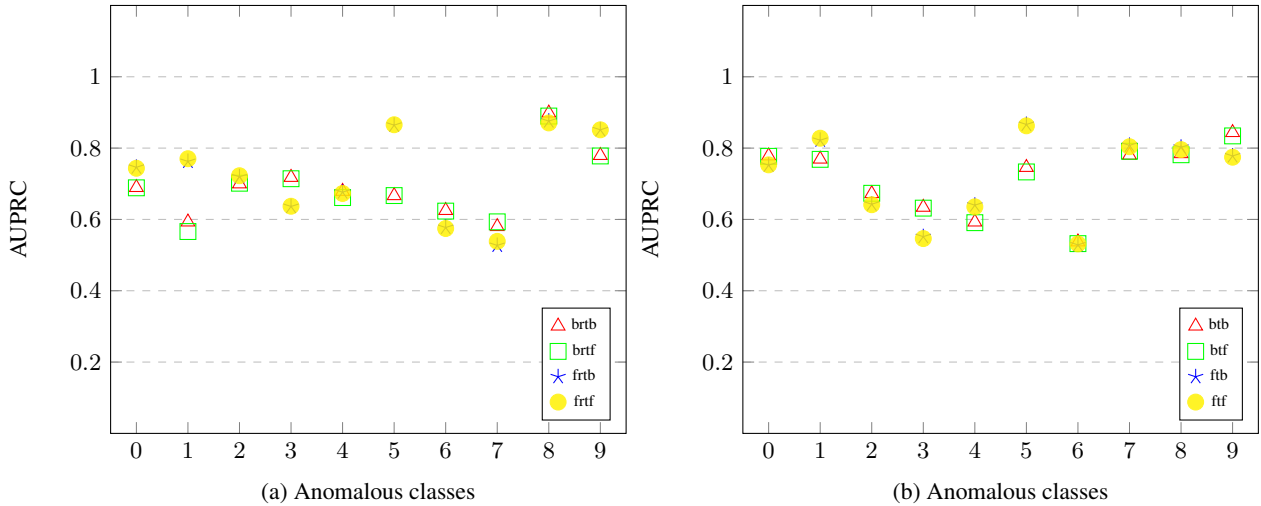
\begin{figure}[!htb]
\centering
{\centering \textbf{Fashion-MNIST Tests (w and w/o residual loss)}\par\medskip}
\begin{tikzpicture}[scale=1,line width=1pt, mark size=3pt]
\begin{axis}[
    xlabel={(a) Anomalous classes},
    ylabel={AUPRC},
    xmin=-0.5, xmax=9.5,
    ymin=0, ymax=1.2,
    xtick={0, 1, 2, 3, 4, 5, 6, 7, 8, 9},
    ytick={0.2, 0.4, 0.6, 0.8, 1},
    legend pos=south east,
    legend style={font=\fontsize{6}{5}\selectfont},
    ymajorgrids=true,
    grid style=dashed,
]
    \addplot[ 
    only marks,
    color=red,
    mark=triangle,
    error bars/.cd, y dir=both,y explicit
    ]
    coordinates {
    (0,0.689448)
	(1,0.592795)
	(2,0.699613)
	(3,0.717808)
	(4,0.678727)
	(5,0.666604)
	(6,0.625060)
	(7,0.579909)
	(8,0.898357)
	(9,0.779284)
    };
    
    \addplot[ 
    only marks,
    color=green,
    mark=square,
    error bars/.cd, y dir=both,y explicit
    ]
    coordinates {
    (0,0.688650)
	(1,0.566025)
	(2,0.701535)
	(3,0.714143)
	(4,0.661283)
	(5,0.666886)
	(6,0.623313)
	(7,0.593230)
	(8,0.889905)
	(9,0.777975)
    };
    
    \addplot[ 
    only marks,
    color=blue,
    mark=star,
    error bars/.cd, y dir=both,y explicit
    ]
    coordinates {
    (0,0.746173)
	(1,0.762642)
	(2,0.719549)
	(3,0.636323)
	(4,0.676430)
	(5,0.864342)
	(6,0.577269)
	(7,0.527153)
	(8,0.875713)
	(9,0.851333)
    };
    
    \addplot[ 
    only marks,
    color=yellow,
    opacity=0.85,
    mark=*,
    error bars/.cd, y dir=both,y explicit
    ]
    coordinates {
    (0,0.743456)
	(1,0.770209)
	(2,0.722502)
	(3,0.636989)
	(4,0.672785)
	(5,0.865463)
	(6,0.574543)
	(7,0.538550)
	(8,0.870631)
	(9,0.851258)
    };
    
    \legend{brtb, brtf, frtb, frtf}
\end{axis}
\end{tikzpicture}
\begin{tikzpicture}[scale=1,line width=1pt, mark size=3pt]
\begin{axis}[
    xlabel={(b) Anomalous classes},
    ylabel={AUPRC},
    xmin=-0.5, xmax=9.5,
    ymin=0, ymax=1.2,
    xtick={0, 1, 2, 3, 4, 5, 6, 7, 8, 9},
    ytick={0.2, 0.4, 0.6, 0.8, 1},
    legend pos=south east,
    legend style={font=\fontsize{6}{5}\selectfont},
    ymajorgrids=true,
    grid style=dashed,
]
    \addplot[ 
    only marks,
    color=red,
    mark=triangle,
    error bars/.cd, y dir=both,y explicit
    ]
    coordinates {
    (0,0.778119)
	(1,0.769129)
	(2,0.672308)
	(3,0.634060)
	(4,0.592735)
	(5,0.745795)
	(6,0.536944)
	(7,0.780044)
	(8,0.784356)
	(9,0.843569)
    };
    
    \addplot[ 
    only marks,
    color=green,
    mark=square,
    error bars/.cd, y dir=both,y explicit
    ]
    coordinates {
    (0,0.776897)
	(1,0.768464)
	(2,0.673277)
	(3,0.631663)
	(4,0.590897)
	(5,0.733163)
	(6,0.532121)
	(7,0.791158)
	(8,0.781356)
	(9,0.834239)
    };
    
    \addplot[ 
    only marks,
    color=blue,
    mark=star,
    error bars/.cd, y dir=both,y explicit
    ]
    coordinates {
    (0,0.752133)
	(1,0.822069)
	(2,0.642627)
	(3,0.551682)
	(4,0.639933)
	(5,0.866280)
	(6,0.528636)
	(7,0.807612)
	(8,0.802364)
	(9,0.777958)
    };
    
    \addplot[ 
    only marks,
    color=yellow,
    opacity=0.85,
    mark=*,
    error bars/.cd, y dir=both,y explicit
    ]
    coordinates {
    (0,0.753034)
	(1,0.827751)
	(2,0.641233)
	(3,0.546273)
	(4,0.635366)
	(5,0.862490)
	(6,0.531388)
	(7,0.803866)
	(8,0.795724)
	(9,0.774468)
    };
    
    \legend{btb, btf, ftb, ftf}
\end{axis}
\end{tikzpicture}
\caption{Performance of version 2 of BiGAN/EGBAD on Fashion-MNIST measured by the area under the precision-recall curve. Best viewed in color. Image (a) depicts the results using the residual loss, image (b) it is without the residual loss. "b(r)tb": bce (+ residual loss) trained and bce tested, "b(r)tf": bce (+ residual loss) trained and fm tested, "f(r)tb": fm (+ residual loss) trained and bce tested, "f(r)tf": fm (+ residual loss) trained and fm tested}
\label{img:BiGAN_fashion_mnist_v2}
\end{figure}

\begin{figure}[!htb]
\centering
{\centering \textbf{GANomaly test results on Fashion-MNIST dataset}\par\medskip}
\begin{tikzpicture}[scale=1,line width=1pt, mark size=3pt]
\begin{axis}[
    xlabel={(a) Anomalous classes},
    ylabel={AUPRC},
    xmin=-0.5, xmax=9.5,
    ymin=0, ymax=1.2,
    xtick={0, 1, 2, 3, 4, 5, 6, 7, 8, 9},
    ytick={0.2, 0.4, 0.6, 0.8, 1},
    legend pos=north west,
    legend style={font=\fontsize{6}{5}\selectfont},
    ymajorgrids=true,
    grid style=dashed,
]   
    \addplot[ 
    only marks,
    color=blue,
    mark=star,
    error bars/.cd, y dir=both,y explicit
    ]
    coordinates {
    (0,0.409281)
	(1,0.649850)
	(2,0.407521)
	(3,0.528104)
	(4,0.432267)
	(5,0.838879)
	(6,0.383896)
	(7,0.501317)
	(8,0.886603)
	(9,0.632766)
    };
    
    \addplot[ 
    only marks,
    color=yellow,
    opacity=0.85,
    mark=*,
    error bars/.cd, y dir=both,y explicit
    ]
    coordinates {
    (0,0.427599)
	(1,0.723826)
	(2,0.377693)
	(3,0.469422)
	(4,0.459695)
	(5,0.804324)
	(6,0.371759)
	(7,0.552956)
	(8,0.879949)
	(9,0.650393)
    };
    \legend{F-MNIST bce, F-MNIST fm}
\end{axis}
\end{tikzpicture}
\caption{GANomaly results on Fashion-MNIST. All tests are here performed through a bce metric and using the area under the precision-recall curve (AURPC). Training is done with a bce loss and a fm loss.}
\label{fig:ganomaly_fashion_mnist_results_v2}
\end{figure}
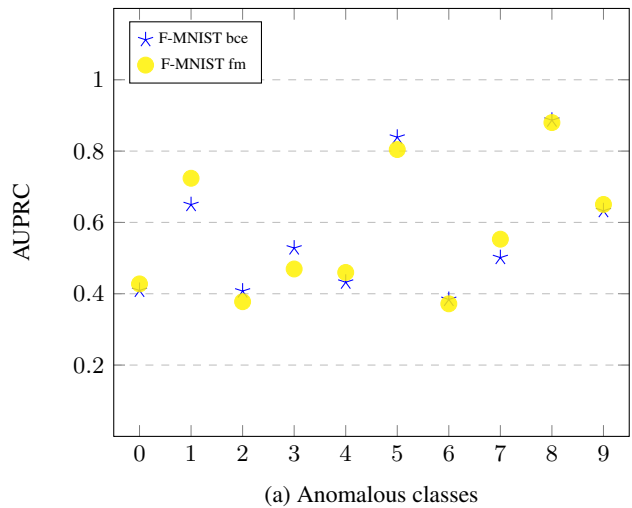

In the following, we present the plots regarding the performances on Fashion-MNIST dataset. \cref{img:BiGAN_fashion_mnist_v2} shows the performance of BiGAN/EGBAD model and \cref{fig:ganomaly_fashion_mnist_results_v2} the performance of GANomaly model.

\clearpage
\clearpage

\subsubsection{CIFAR10}
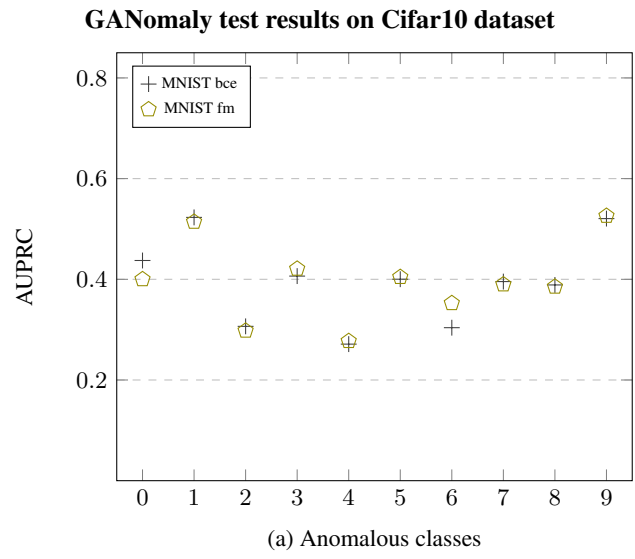
\begin{figure}[!htb]
\centering
{\centering \textbf{GANomaly test results on Cifar10 dataset}\par\medskip}
\begin{tikzpicture}[scale=1,line width=1pt, mark size=3pt]
\begin{axis}[
    xlabel={(a) Anomalous classes},
    ylabel={AUPRC},
    xmin=-0.5, xmax=9.5,
    ymin=0, ymax=0.85,
    xtick={0, 1, 2, 3, 4, 5, 6, 7, 8, 9},
    ytick={0.2, 0.4, 0.6, 0.8, 1},
    legend pos=north west,
    legend style={font=\fontsize{6}{5}\selectfont},
    ymajorgrids=true,
    grid style=dashed,
]   
    \addplot[ 
    only marks,
    color=darkgray,
    mark=+,
    error bars/.cd, y dir=both,y explicit
    ]
    coordinates {
	(0,0.437373)
	(1,0.522991)
	(2,0.306560)
	(3,0.406376)
	(4,0.271251)
	(5,0.400210)
	(6,0.303915)
	(7,0.395485)
	(8,0.388809)
	(9,0.520576)
    };
    
    \addplot[ 
    only marks,
    color=olive,
    mark=pentagon,
    error bars/.cd, y dir=both,y explicit
    ]
    coordinates {
	(0,0.400025)
	(1,0.514118)
	(2,0.297837)
	(3,0.420892)
	(4,0.277245)
	(5,0.404456)
	(6,0.352527)
	(7,0.389732)
	(8,0.385233)
	(9,0.525924)
    };
    
    \legend{MNIST bce, MNIST fm, F-MNIST bce, F-MNIST fm, Cifar10 bce, Cifar10 fm}
\end{axis}
\end{tikzpicture}
\caption{GANomaly results on CIFAR10. All tests are here performed through a bce metric and using the area under the precision-recall curve (AURPC). Training is done with a bce loss and a fm loss.}
\label{fig:ganomaly_cifar10_results_v2}
\end{figure}
\clearpage

\end{document}